\documentclass[conference]{IEEEtran}
\IEEEoverridecommandlockouts
\usepackage{cite}
\usepackage{amsmath,amssymb,amsfonts}
\usepackage{graphicx}
\usepackage{textcomp}
\usepackage{xcolor}

\usepackage{amsmath,graphicx}
\usepackage{booktabs}
\usepackage{hyperref}

\usepackage[ruled,vlined]{algorithm2e}
\usepackage{amsmath}
\usepackage{wrapfig}
\usepackage{multirow}
\usepackage{tabularx}
\usepackage{graphicx}
\usepackage{adjustbox}
\usepackage{amsmath}

\usepackage[english]{babel}
\usepackage[utf8]{inputenc}
\usepackage[noend]{algpseudocode}

\def\BibTeX{{\rm B\kern-.05em{\sc i\kern-.025em b}\kern-.08em
    T\kern-.1667em\lower.7ex\hbox{E}\kern-.125emX}}
\begin{document}

\title{HyperGALE: ASD Classification via \underline{Hyper}graph \underline{G}ated \underline{A}ttention with \underline{L}earnable Hyper\underline{E}dges
}
% Mehul Arora, Chirag S. Jain, Lalith Bharadwaj Baru, Kamalaker Dadi,  Raju S. Bapi

% \author{\IEEEauthorblockN{Authors}
% \IEEEauthorblockA{{xxxxx} \\
% \textit{xxxxxxxxxx}\\
% % City, Country \\
% % {mehul.arora, chirag.jain}@research.iiit.ac.in
% }}

\author{\IEEEauthorblockN{Mehul Arora, Chirag S. Jain, Lalith Bharadwaj Baru, Kamalaker Dadi, Bapi Raju S}
\IEEEauthorblockA{{Brain Cognition Computation Lab} \\
\textit{IIIT Hyderabad, India}\\
% City, Country \\
{mehul.arora, chirag.jain}@research.iiit.ac.in}
}

\maketitle

\begin{abstract}
Autism Spectrum Disorder (ASD) is a neurodevelopmental condition characterized by varied social cognitive challenges and repetitive behavioral patterns. Identifying reliable brain imaging-based biomarkers for ASD has been a persistent challenge due to the spectrum's diverse symptomatology. Existing baselines in the field have made significant strides in this direction, yet there remains room for improvement in both performance and interpretability. We propose \emph{HyperGALE}, which builds upon the hypergraph by incorporating learned hyperedges and gated attention mechanisms. This approach has led to substantial improvements in the model's ability to interpret complex brain graph data, offering deeper insights into ASD biomarker characterization. Evaluated on the extensive ABIDE II dataset, \emph{HyperGALE} not only improves interpretability but also demonstrates statistically significant enhancements in key performance metrics compared to both previous baselines and the foundational hypergraph model. The advancement \emph{HyperGALE} brings to ASD research highlights the potential of sophisticated graph-based techniques in neurodevelopmental studies. The source code and implementation instructions are available at {\color{blue}\href{https://github.com/mehular0ra/HyperGALE}{Github}}.
\end{abstract}
% {{\color{blue}\url{https://github.com/mehular0ra/HyperGALE}}.


\section{Introduction}\label{sec:intro}

\begin{figure*}[!t]
    \centering
    \includegraphics[width=0.86\linewidth]{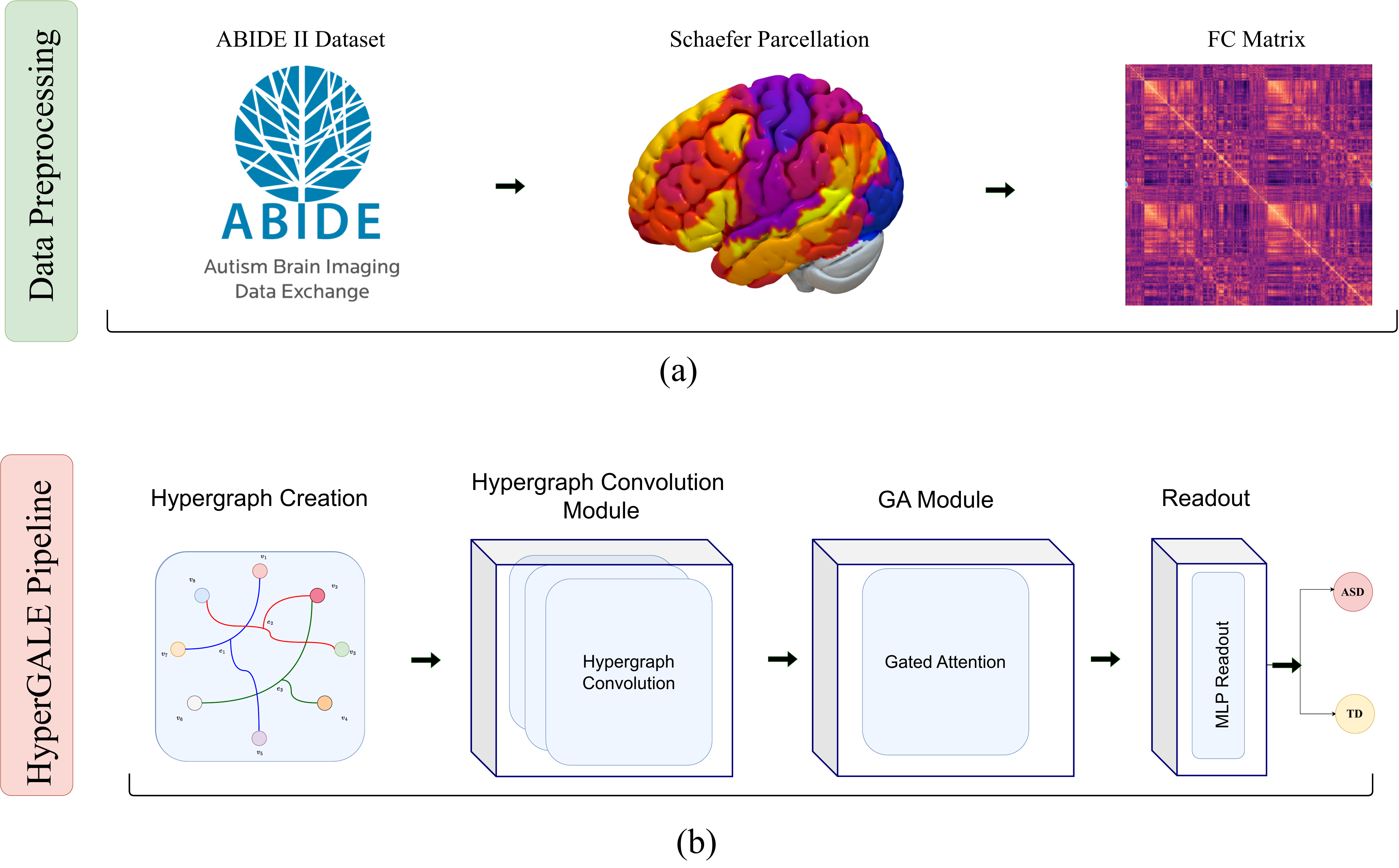}
    \caption{\textbf{ASD Processing Pipeline and the Proposed HyperGale Architecuture}. {(\bf a)} fMRI data is preprocessed with Schaefer's parcellation and later converted into a functional connectivity (FC) matrix. {\bf (b)} The FC matrix is then converted into a hypergraph with learnable hyperedge weights. Subsequently, the hypergraph features are sent to a series of hypergraph convolution layers and gated attention is applied in order to capture the importance of relevant information. Finally, the information extracted after attention layers is aggregated using the readout function. Finally, from these lower dimensional features of readout, a sigmoid activation leads to binary classification of ASD vs Typically Developing (TD).}
    \label{fig-main}
\end{figure*}

Autism Spectrum Disorder (ASD) is a complex neurodevelopmental condition characterized by challenges in social cognition and repetitive behaviors. Affecting a significant portion of the population, with a prevalence of approximately 1 in 36 children \cite{maenner2023prevalence}, ASD's diagnosis is complicated by its heterogeneous nature and the limitations of existing diagnostic criteria. Emerging research suggests that the atypical behaviors observed in ASD may be linked to distinct patterns in functional connectivity (FC) within the brain \cite{uddin2013reconceptualizing}\cite{nomi2015developmental}.

Functional connectivity
%unlike static structural brain connectivity, 
represents dynamic interactions and synchronizations between various brain regions. These interactions form intricate networks that are foundational for cognitive processes and behaviors, exemplified by networks such as the Default Mode Network (DMN) and Dorsal Attention Network (DAN) \cite{smith2009correspondence}. The importance of these networks in ASD is increasingly recognized in contemporary research \cite{bathelt2022more}, underscoring the need for analytical models capable of discerning complex FC patterns associated with ASD.

Several models have been applied to understand ASD, ranging from traditional methods like SVM and Random Forest \cite{arbabshirani2017single} to neural networks, CNNs and Transformers \cite{Chen2024, eslami2019asd, kawahara2017brainnetcnn, vaswani2017attention, kan2022brain}. Graph-based methods have also been explored, retaining complex brain information for ASD classification. However, these methods often overlook higher-order relationships in the brain's network, focusing on dyadic interactions. This limitation is significant in the context of ASD, where the implicated brain regions can be highly variable and diverse \cite{muller2003abnormal}.

% Addressing these challenges, our research introduces \emph{HyperGALE}, a model that employs hypergraph convolutions with gated attention to capture high-order graphical intricacies in the brain. \emph{HyperGALE} stands at the intersection of computational neuroscience and graph theory, offering the following key contributions:

% \begin{enumerate}
% \item The development of a novel approach for ASD classification, leveraging hypergraph convolutions and gated attention to effectively identify specific brain regions associated with ASD, showcasing superior results compared to current state-of-the-art methods.
% \item A rigorous exploration of various hyperparameters, alongside robustness tests including multiple initializations and leave-one-site-out strategies, to establish the model's consistent performance and generalization capabilities across diverse datasets.
% \item Bridging computational analysis with neuroscience, our model offers not just methodological improvements but also valuable interpretative insights into the qualitative aspects of ASD.
% \end{enumerate}

Addressing the intricate challenges in ASD diagnosis, our research introduces \emph{HyperGALE}, at the confluence of computational neuroscience and graph theory. Unlike traditional graph-based methods that primarily focus on pairwise interactions, \emph{HyperGALE} utilizes hypergraph convolutions. This approach allows the model to capture high-order relationships within the brain's network, crucial for understanding the complex and heterogeneous nature of ASD. Coupled with gated attention, \emph{HyperGALE} not only discerns intricate patterns in functional connectivity but also provides a complex understanding of brain region interactions associated with ASD. Our key contributions are as follows:

\begin{enumerate}
\item We developed \emph{HyperGALE}, a novel ASD classification approach utilizing modified hypergraph convolution and gated attention to identify critical brain regions implicated in ASD. Additionally, we re-implemented and compared several holistic baseline models on the ABIDE-II dataset. These comparisons are based on accuracy, AUC, sensitivity, and specificity, highlighting \emph{HyperGALE}'s superior performance.
\item Our investigation into the model's hyperparameters, such as the number of Regions of Interest (ROIs) in hyperedges and the hypergraph layer count, provided insights into their impact on \emph{HyperGALE}'s performance. We also assessed the robustness of all models, including \emph{HyperGALE}, through multiple runs with different initializations and dataset distributions, noting \emph{HyperGALE}'s consistently low standard deviation in outcomes. The model's generalization capabilities were further evidenced by employing a leave-one-site-out strategy.

\item The advancements in \emph{HyperGALE} go beyond performance metrics, offering interpretative insights into ASD's qualitative aspects through learnt hyperedges and gated attention. This bridges the gap between computational analysis and neuroscience, providing a deeper understanding of neurodevelopmental disorders such as ASD.

\end{enumerate}

\section{Related Works}

In this section, we begin by briefly reviewing traditional machine-learning, deep-learning architectures that do not incorporate graph structures. We then transition to discussing the evolution and significance of graph-based and hypergraph-based methods in the context of ASD classification.

\begin{figure*}[!t]
    \centering
    \includegraphics[width=0.90\linewidth]{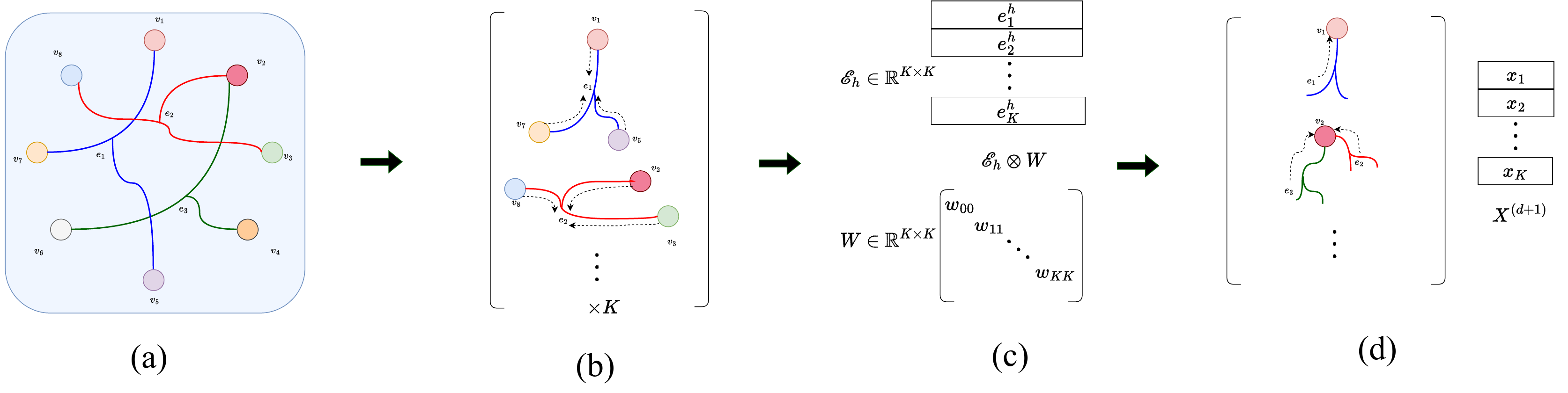}
    \caption{\textbf{ Modified hypergraph convolution proposed in HyperGALE pipeline}. Starting with ROI features derived from the fully connected matrix, these initial inputs undergo node-to-hyperedge propagation to form hyperedge features (step (b)). These features are scaled using learned weights, as in step (c), the activation happens in different hyperedges with different values. Finally, the scaled hyperedge features are propagated to new node features (step (d)).}
    \label{fig:hypergale-pipeline}
\end{figure*}

\subsection{Overview of Non-Graph Methods}
Traditional machine learning methods, such as Support Vector Machines (SVM),
% \cite{cortes1995support}
Random Forests, and Gradient Boosting Classifiers (GBC)\cite{arbabshirani2017single}, have been applied to ASD classification. These methods, however, often face challenges in handling the high variance inherent in ASD data, attributed to site-specific variations and the heterogeneous nature of the disorder.

In the realm of deep learning, architectures like CNNs and Transformers have shown promise. Eslami \emph{et al.} \cite{eslami2019asd} implemented an autoencoder with data augmentation, while Kawahara \emph{et al.} \cite{kawahara2017brainnetcnn} utilized a CNN, treating the adjacency matrix as an image rather than exploiting its inherent graph structure. Kan \emph{et al.}~\cite{kan2022brain} innovatively employed a transformer-based approach for node embeddings with self-supervised clustering for readout.

\subsection{Graph-based Methods}
A myriad of graph-based models was proposed on top of the standard GNNs 
%which led to improvements in this field 
\cite{kipf2016semi}, \cite{velivckovic2017graph}, \cite{hamilton2017inductive}. For example, Cao \emph{et al.} \cite{cao2021using} constructed a deep ASD diagnostic framework based on 16-layer GCN with ResNet units and DropEdge strategy to avoid certain problems such as vanishing gradient, and over-smoothing. Kazi \emph{et al.} \cite{kazi2019inceptiongcn} proposes `inception modules’ which are capable of capturing intra- and inter-graph structural heterogeneity during convolutions. Yao \emph{et al.} \cite{yao2019triplet} introduce a multi-scale triplet graph convolutional network that employs multi-scale templates and a triplet GCN (TGCN) model to learn multi-scale graph representations of brain FC. A recent significant study by Chen \emph{et al.} \cite{Chen2024} put forth a competitive model that used both the resting state fMRI (brain function) and T1 weighted MRI (brain structure) to generate separate node and edge embeddings using a Transformer block.

Since the FC matrix is symmetric, traditional models tend to use the upper/lower triangular matrix features of FC, thereby lacking message passing. Therefore, these methods struggle to aggregate local and global information, resulting in poor performance. The CNNs benefit from translational equivariance and the Transformer's key feature is the self-attention mechanism. Although, these methods have good properties, due to a lack of message passing they fail to capture aggregated neighbourhood information. Graph-based approaches on FC matrices are applied to understand interactions among regions of interest, but current methods can not take advantage of higher-order proximity information due to reliance on static dyadic edges. To address these limitations and accommodate the heterogeneity in individuals with ASD, exploring hypergraph models is considered to be a promising step.

\subsection{Hypergraph-based Methods}
Recent ASD classification studies have explored hypergraph frameworks, such as Hypergraph U-Net\cite{lostar2020deep} and multi-view HGNN\cite{madine2020diagnosing}, utilizing limited multi-modal data. Frameworks like those of Shao \emph{et al.} \cite{shao2020hypergraph} for Alzheimer's classification incorporate feature selection and group-sparsity regularization in their hypergraph approach. Similarly, Liu \emph{et al.} \cite{liu2017view} use a view-aligned regularizer for multimodal coherence in hypergraph creation. Another notable work, \cite{ji2022fc}, employs dynamic hypergraph neural networks, utilizing KNN and KMeans for iterative hypergraph generation in ASD classification.

Different from these methods, \emph{HyperGALE} utilizes the brain's inherent hypergraph structure with a specialized hypergraph convolution technique and learnable hyperedges. This approach enhances the interpretation of complex brain networks for more effective ASD classification.

\section{Methodology}
In this section, we initially discuss each component of our method step-by-step and eventually detail the entire pipeline. 

\subsection{Dataset and Preprocessing}
To systematically benchmark various graph learning techniques for fMRI-based disease classification, we used large-scale openly-accessible fMRI dataset from ABIDE-II consortium\footnote{\url{http://fcon_1000.projects.nitrc.org/indi/abide/abide\_II.html}}. This work included 812 subjects, out of which ASD: 384 and TD: 428, respectively, accumulated across 16 sites as shown on the left of Figure~\ref{fig:leave-one-site-out}. The resting-state fMRI dataset was pre-processed using the default processing pipeline considering global signal regression and bandpass filtering implemented with C-PAC pipeline \cite{cpac2013}.

Briefly, the pre-processing pipeline included: removing the skull regions and segmentation of each anatomical image into three tissues, followed by normalization to the MNI152 template using Advanced Normalization Tools (ANTs)~\cite{avants2011}. fMRI pre-processing included slice timing correction, motion correction, global mean intensity normalization as well as nuisance signal regression. In nuisance signal regression, the number of parameters used were 24 parameters for head motion, CompCor~\cite{behzadi2007} with five principal components signals from Cerebrospinal fluid and white matter, linear and quadratic trends for low-frequency drifts. fMRI images were then co-registered with subject-specific corresponding anatomical images and finally normalized to the MNI152 space using ANTs.

After pre-processing, timeseries signals were extracted for each node where each node belongs to 7 networks 400 parcellations from Schaefer atlas~\cite{schaefer2018}. On each node time series, functional connectivity estimation using the Ledoit-Wolf regularized shrinkage estimator~\cite{ledoit2004} and full correlation was implemented with Nilearn package~\cite{abraham2014}.

\subsection{Hypergraph Modeling}
As introduced earlier, we construct a graph based on the FC correlation matrix. Also, the fundamental notations and the formulations are adopted from Feng \emph{et al.} \cite{feng2019hypergraph}.
%---I don't think this is good
% Here, each node of a graph corresponds to a unique brain region. Here, we first describe the basic graph mechanism and later introduce hypergraphs. Here the fundamental notations and the formulations are adopted from the research of Bai \emph{et al.} \cite{bai2021hypergraph} and Feng \emph{et al.} \cite{feng2019hypergraph}.   
\textbf{Basic Notations:}
Suppose $\mathcal{G} = (\mathcal{V}, \mathcal{E})$ be our graph with $\mathcal{V}$ as vertices and $\mathcal{E}$ edges respectively. Where, $\mathcal{V} \in \{v_1, v_2, \cdots v_N \}$ and $\mathcal{E} \subseteq \mathcal{V} \times \mathcal{V}$. The Adjacency matrix $\mathbb{A} \in \mathbb{R}^{N \times N}$ informs us of the pairwise interactions between each and every node.

Similarly, to construct a hypergraph $\mathcal{G}_{hyper} = (\mathcal{V}, \mathcal{E}_h)$ with $N$ nodes and $K$ hyperedges. Similar to the adjacency matrix in the graph, in hypergraph we have \emph{Incidence matrix} $\mathcal{H} \in \mathbb{R}^{N \times K}$. Here, the hyperedges $e_j \in \mathcal{E}_h $ ($j=1, 2, \cdots K$) are dumped in a diagonal weight matrix $W \in \mathbb{R}^{K \times K}$. The entries in the aforementioned incidence matrix can be represented as 
$$\mathcal{H}_{ij} = 
\begin{cases}
    1 &\quad \text{if } v_i \in e_j\\
    0 &\quad\text{Else.} \\
\end{cases}$$

The vertex degree and the hyperedge degree of the hypergraph are defined as $\mathcal{D}_{ii} = \sum_{j=1}^K W_{jj} \mathcal{H}_{ij}$ and $\mathcal{B}_{jj} = \sum_{i=1}^N \mathcal{H}_{ij}$ respectively. 

\begin{table*}[t!]
\centering
\caption{The Table illustrates the performance of our proposed method HyperGALE as compared to the existing state-of-the-art approaches. The train and test proportions considered for the study are 90\% and 10\%, respectively. The experiments are conducted for 5 different non-overlapping proportions of data and the mean and standard deviations are reported. }\label{tab-performance}
\scalebox{0.95}{
\begin{tabular}{ c|l|cccc } 
\hline
\multirow{2}{8em}{Modality} & \multirow{2}{5em}{\textbf{Methods}} & \multicolumn{4}{c}{\textbf{Performance Metrics}} \\ \cline{3-6}
   & & Accuracy & AUC & Sensitivity & Specificity \\ \hline \hline
  
  \multirow{3}{8em}{\textbf{Traditional Methods}}  
  &SVM \cite{cortes1995support} & 68.83$\pm$ 2.44	&68.07$\pm$ 2.49	&78.82$\pm$ 2.65	&57.31$\pm$ 3.01   \\
  
  &Random Forest \cite{breiman2001random}&60.00$\pm${ 5.55}&	58.80$\pm${6.11}&	70.81$\pm${ 4.11}	&48.90$\pm${7.24} \\
  
  & Gradient Boosting  &62.01$\pm$ 2.99	&61.45$\pm$ 2.92	&68.86$\pm$ 4.54	&54.05$\pm$ 5.23\\ \hline

  \multirow{2}{8em}{\textbf{Non-graph  Methods}}  
  &ASD-Diagnet  \cite{eslami2019asd} &70.8 $\pm${1.49} &71.12 $\pm${2.06} &68.36 $\pm${8.13} &72.2 $\pm${10.56} \\	  
  
  &BrainNetCNN \cite{kawahara2017brainnetcnn}& 67.32$\pm$ 4.24&	69.07$\pm$ 3.05&	68.89$\pm$ 8.62&	65.40$\pm$ 11.56\\  \hline
  
  % & METHOD3 & & & & & \hline

  \multirow{3}{8em}{\textbf{Graph based Methods}}  
  &GCN  \cite{kipf2016semi}&72.68$\pm${3.49}	&75.65$\pm${ 3.57}	&\bf 79.56$\pm${ \bf 8.42}	&64.33$\pm${ 9.65} 	\\  
  
  &GAT  \cite{velivckovic2017graph}&68.53$\pm${ 3.40}	&71.70$\pm${ 4.11}	&65.78$\pm${ 9.76}	&71.89$\pm${ 9.75} \\
  
  &GraphSAGE \cite{hamilton2017inductive} &71.22$\pm${2.55}	&74.64$\pm${4.60}	&75.43$\pm${17.88}	&66.61$\pm${2.56} \\\hline

  \multirow{2}{8em}{\textbf{Transformer Based Methods}}  
  &Transformer \cite{vaswani2017attention} & 67.80$\pm$ 0.98&	69.14$\pm$ 1.87&	62.99$\pm$ 6.55&	72.50$\pm$ 4.07 \\  
  
  &BrainNetTransformer \cite{kan2022brain}&
  71.22$\pm$ 0.97&	73.60$\pm$ 2.38&	67.70$\pm$ 8.27&	\bf 74.29$\pm$ \bf 5.54   \\\hline
  
  % & METHOD 3 & & & & & \hline

  \multirow{2}{8em}{\textbf{HyperGraphs (Ours)}}  
  &HyperGraphGCN %\cite{feng2019hypergraph} 
  & 73.41$\pm${1.18}	&76.66$\pm${1.02}	&74.67$\pm${5.76}	&71.89$\pm${5.91}	  \\
  
  % &HyperGraphGCN v2.0 &  73.74$\pm${1.39}  & 77.15$\pm${3.80} &68.10$\pm${7.20}	&\bf 81.15$\pm${\bf 6.56} &  
  &HyperGALE (Ours) & \bf 75.34$\pm$ \bf 0.47	& \bf 77.03$\pm$ \bf 1.85 &	 76.39$\pm$  4.84	& 73.91$\pm$  5.58 \\ \hline
  
  % & METHOD 3 & & & & & 
  % \hline
\end{tabular}
}
\end{table*}

\textbf{Hypergraph Convolutions:}
Now, the incidence matrix $\mathcal{H}$ is passed to a successive series of convolution layers. In order for the convolution operation to proceed the propagation is done among the shared hyperedges and higher confidence is assigned to the larger weights. Based on the hypergraph structure and the associated weights, we define the hypergraph convolution operation as,

\begin{equation}\label{eq-hypergraph-nonmatrix}
    x_t^{(d+1)} = \sigma \left( \sum_{i=1}^N \sum_{j=1}^K \mathcal{H}_{tj} \mathcal{H}_{ij} W_{jj} x_t^{(d)} \Theta  \right)
\end{equation}

Here $x_t^{(d)}$ is the embedding of the $t^\text{th}$ vertex in the $d^\text{th}$ layer. The $\Theta$ is the weight matrix between two successive layers. Now we can eventually formulate the equation (\ref{eq-hypergraph-nonmatrix}) into a matrix form as 

\begin{equation}\label{eq-hypergraph-matrix}
    X^{(d+1)} = \sigma \left( \mathcal{H} W  \mathcal{H}^T X^{(d)} \Theta  \right)
\end{equation}
From the above equation the $X^{(d)} \in \mathbb{R}^{N \times F^{(d)}}$ is the input of the $d^\text{th}$ layer and the $\Theta \in \mathbb{R}^{F^{(d)} \times F^{(d+1)} }$. The equation (\ref{eq-hypergraph-matrix}) refers to a basic formulation that might lead to numerical instabilities with deepening layers causing gradients to vanish. Thus, to tackle this, one can adopt normalization approaches, such as symmetric normalization or row-normalization. In our line of research, we adopt the row-normalization approach as it provides directional propagation as in the below equation 

\begin{equation}\label{eq-hypergraph-normmatrix}
    X^{(d+1)} = \sigma \left( \mathcal{D}^{-1}\mathcal{H} W \mathcal{B}^{-1} \mathcal{H}^T X^{(d)} \Theta  \right)
\end{equation}

% \paragraph{Definitions}
\textbf{Learnable Hyperedges and Gated Attention}
In the equation (\ref{eq-hypergraph-normmatrix}) the weight matrix is not learnable. In such cases, the hypergraph will not be updated with respect to the samples. So our method even tries to learn the hyperedge weights matrix $W$ and we represent it as $\Tilde{W}$ (refer to fig. \ref{fig:hypergale-pipeline}). Successively, this information is passed through a gated attention network as formulated below,  
\begin{align}\label{eq-hypergraph-gated-attention} \notag
    A  = \sigma ( \textsc{Mlp}(X^{(d+1)}));\\ 
    B = \text{Tanh} ( \textsc{Mlp}(X^{(d+1)}));\\ \notag
    \alpha = \textsc{MLP}(A \odot B); \\
    Z = \alpha  \odot X^{(d+1)}\notag
\end{align}

\begin{figure}[!t]
    \centering
    \includegraphics[width=1.0\linewidth]{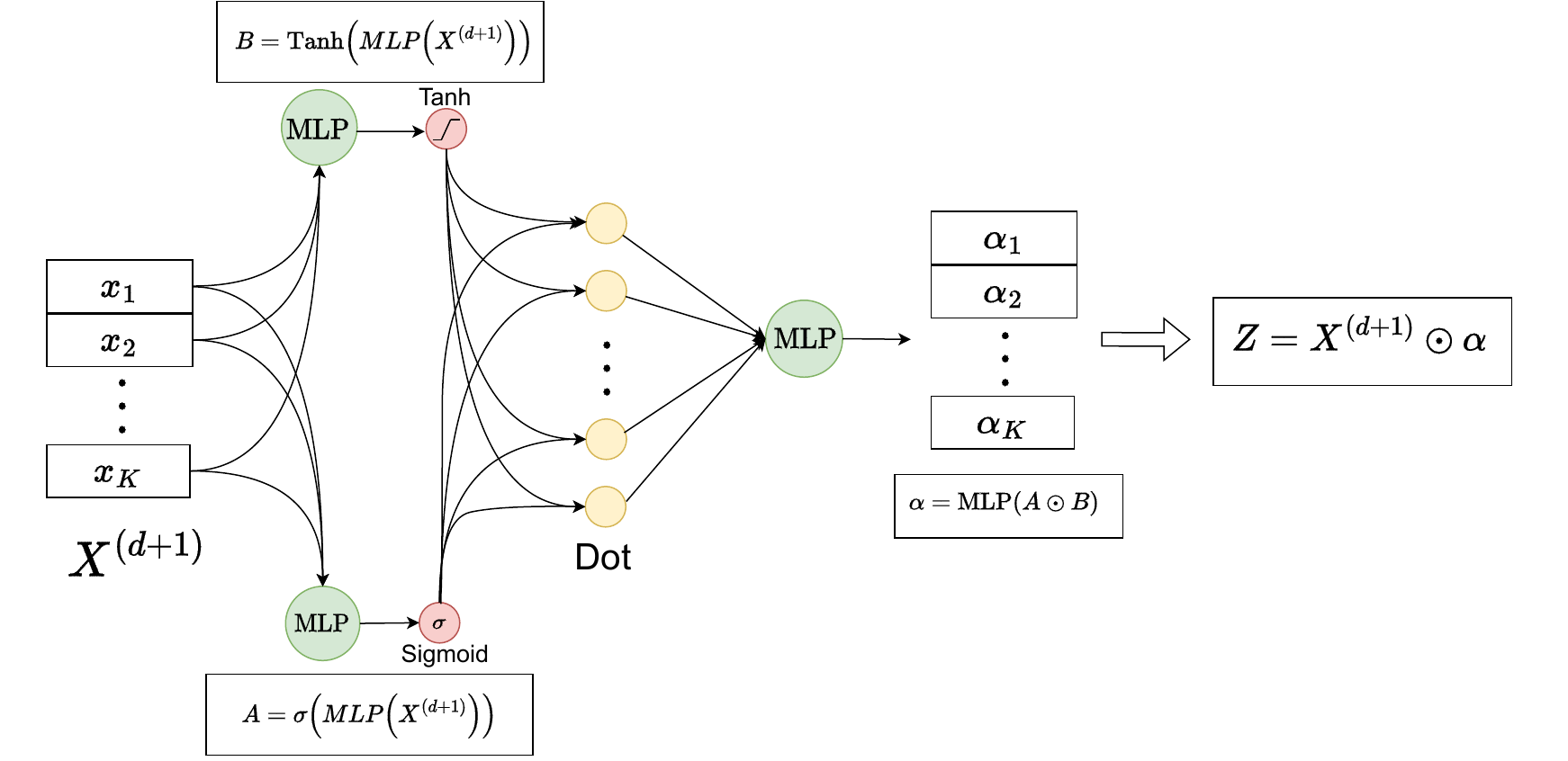}
    \caption{\textbf{ Illustration of the Gated Attention Module}. GA module which learns $\alpha$ iteratively (or numerically) from eq. (\ref{eq-hypergraph-gated-attention}), this $\alpha$ vector is multiplied to get the final node features which is fed to the readout layer. \cite{ilse2018attention}.}   
    \label{fig:gated-attention}
\end{figure}

Where, $\odot$ determines the element-wise product and $\text{Tanh}(\cdot)$, $\sigma (\cdot)$ are the nonlinear activation's respectively. In the equation (\ref{eq-hypergraph-gated-attention}), $Z$ obliges our model to allocate necessary attention that are crucial to interpreting the exact incidence of ASD \cite{ilse2018attention}. This guides the hyperedge weight matrix to direct the importance to the essential nodes. Thus, it eventually tracks essential regions of the brain that are actually connected and injects the importance accordingly (refer to Fig. \ref{fig:gated-attention}).

\textbf{Readout Mechanisms} The Readout methods produce an aggregation of node-level features to form graph-level representations. The standard readouts such as, $\textsc{max} (\cdot) $ and $\textsc{mean} (\cdot)$ underfit while learning on brain hypergraphs. So we experimented with adaptive readouts like $\textsc{SetTransformer}$ and $\textsc{Janossy}$ (permutation invariant) and MLP (permutation non-invariant). Thus, we deploy \emph{Neural Readouts} where the hypothesis space of the aggregated representations is not permutation invariant. Also similar to Buterez \emph{et al.} \cite{buterez2022graph} we applied three variants of adaptive neural readouts, $\textsc{SetTransformer}$, $\textsc{Janossy}$ and $\textsc{Mlp}$ readouts. We experimented with both permutation variant and invariant that are adaptive and embedded with permutation invariant learning with and without attention mechanisms \cite{buterez2022graph}. After a detailed experiments we have seen $\textsc{Mlp}$ readouts outperform many adaptive and non-adaptive readouts. The ablations are detailed briefly in the results section. Our chosen readout is,
\begin{equation}
   X^\text{(Readout)} = \textsc{Relu}( \textsc{Mlp}(Z))
\end{equation}
Finally, the total pipeline of our HyperGALE is described in the below Algorithm.{\color{blue}\ref{alg:HyperGALE}}

% \subsection{Complete Pipeline} 
% % FC -> Hypergraph Generation -> Hypergraph convolution + Hyperedge weight learning -> MLP Readout -> (Prediction)
% Here we summarise the complete pipeline used to produce the desired classification of subjects with the Algorithm \ref{alg:HyperGALE}. This Algorithm describes the construction of the hypergraphs with a preprocessed FC matrix given as input resulting in the classification of ASD or TD (Typically Developing). The methodological diagram and individual components corresponding to 'HyperGALE' are elucidated in Fig. \ref{fig-main}. For visualization, we relied on Nilearn \cite{abraham2014}.
%%%%---------------Algo: HyperGALE-------------%%%%%
\begin{algorithm}
\caption{$\textsc{HyperGALE}$ }\label{alg:HyperGALE}
% \SetAlgoNlRelativeSize{0}
% \SetNlSty{textbf}{(}{)}
% \SetAlgoNlRelativeSize{1}
% \SetAlgoNlRelativeSize{-1}
% \DontPrintSemicolon
\KwIn{\textsc{Data}}
\KwOut{$\hat{y}$}
% \BlankLine\SetKwData{\mathcal{D}}
% {\mathcal{D}}\SetKwData{K}{k}
% \SetKwData{Hypergraph}{H}
% \SetKwData{Node}{i}
% \SetKwData{Neighbors}{neighbors}\SetKwData{Idx}{j}\SetKwData{Neighbor}{neighbor}\BlankLine

$FC \leftarrow \textsc{SchaeferParcellation} (\textsc{Data}) $

\For{i $\leftarrow$ 1 \KwTo epochs}
{
$\;\; \text{Hypergraph Creation: }\mathcal{G}_{hyper} (V, \mathcal{H})$
$X^{(d+1)} \leftarrow \sigma \left( \mathcal{D}^{-1}\mathcal{H} \Tilde{W} \mathcal{B}^{-1} \mathcal{H}^T X^{(d)} \Theta  \right)$
$\;\;Z \leftarrow  \alpha \odot X^{(d+1)}$

$\;\;X^\text{(Readout)} \leftarrow \textsc{Relu}(\textsc{Mlp}(Z))$

$\;\;\hat{y} \leftarrow \sigma(X^\text{(Readout)})$
}
\KwRet{$\hat{y}$}
\end{algorithm}

% Now, this $\textsc{HyperG}$ is incorporated in our $\textsc{HyperGALE}$ model. In the Algorithm \ref{alg:HyperGALE} the $\textsc{Data}$ refers to the preprocessed FC matrix. Thus each step is similar to the HyperGraphCNN but the major difference is we make the weights of the hyperedges learnable to align with the subject data resulting in an informative outcome.
% % HyperGALE Algorithm is as follows

\section{Results and Discussion}
In this section, we compare our proposed method with methods that are employed to provide significant performance for ASD classification. Later we determine the qualitative performance of HyperGALE in determining ASD. 

\subsection{Methods for Comparison}
All the baseline experiments have been conducted on the dataset we have identified from ABIDE II and where necessary we have re-implemented the algorithms for reporting the comparative results in Table~\ref{tab-performance}.
First, to observe the significance of statistical approaches the traditional machine learning methods such as Support Vector Machines, Random Forests, and Gradient Boosting Classifier (GBC) %\cite{friedman2001greedy}
have been implemented. These models have been poorly performed due to the high variance nature of the data (refer Table \ref{tab-performance}). Nevertheless, comparing among traditional methods, SVM gave better prediction performance.
%\cite{abraham2017deriving}. 
% These traditional methods were observed from Chen \emph{et al.}~\cite{Chen2024} but, we re-implemented them for our study as they differ in parcellation and subjects considered for the study.  

% % Input to these models is the upper triangular part of the functional connectivity matrix. 
%  Gradient Boosting Classifier, being an ensemble method that corrects for residuals (errors) from previous models was able to perform well.

% \subsection{Non-Graph Methods}
Next, moving to the architectures using deep learning without transformer, the works \cite{eslami2019asd, kawahara2017brainnetcnn} have decent and better performance compared to traditional methods. Although, these methods have succeeded to provide jump in performance compared to that of traditonal methods these models have high standard deviation and needs to be better in capturing the patterns that corresponds to ASD.  

Next, we move to latest transformer architectures \cite{vaswani2017attention, kan2022brain} have achieved promising performance with less standard deviation in AUC and accuracy metrics. Although, having self-attention layers and OCReadout mechansims, these methods do not provide significant performance. 

Now, we benchmark the graph-based models \cite{kipf2016semi, velivckovic2017graph, hamilton2017inductive}. As explained earlier, graph-based have actually performed well for ASD dataset \cite{cao2021using}\cite{kazi2019inceptiongcn} \cite{yao2019triplet}. Graph-based approaches on FC matrices are applied to understand interactions among regions of interest, but current methods struggle with higher-order proximity due to reliance on static dyadic edges.

% \subsection{Hypergraph-based Methods}
Recent studies using Hypergraphs for ASD classification include Hypergraph U-Net \cite{lostar2020deep} and multi-view HGNN\cite{madine2020diagnosing} which use multiple modalities and thus we can't employ these methods for our study. Finally, we employed the hypergraph convolution methods named HyperGraphGCN \cite{feng2019hypergraph} to compare with the hypergraph baseline. Although these methods have the added advantage of acquiring higher-order proximity, HyperGALE performs extensively on top of every model. In the next subsection, we explain all the experiments in detail and justify HyperGALE with extensive ablation studies. 

% % \cite{shao2020hypergraph} and \cite{liu2017view} are another set of hypergraph frameworks for Alzhiemer classification. Shao \emph{et al.} \cite{shao2020hypergraph} perform feature selection on each modality followed by group-sparsity regularizer which are given to the Hypergraph and finally use SVM for classification. Liu \emph{et al.} \cite{liu2017view} perform multimodality study which uses view-aligned regularizer to capture coherence among views which are further used to create hypergraphs.
% \cite{ji2022fc} uses dynamic hypergraph neural networks for ASD classification which involves k-NN and k-means for hypergraph generation in every iteration followed by hypergraph aggregation which is computationally expensive and do not produce significant results.  due \emph{attention} and \emph{learnability} of hyperedges.   

\begin{table}[t!]
\centering
\caption{The Table illustrates the performance hypergraph models in the combinations of with and without GA and learned HE.}\label{tab-GA-HE}
\scalebox{0.90}{
\begin{tabular}{ l|l|cccc } 
\hline
\multicolumn{2}{c}{\textbf{Methods}} & \multicolumn{4}{c}{\textbf{Performance Metrics}} \\ \cline{1-6}
    Learned HEs & GA & Accuracy & AUC & Sensitivity & Specificity \\ \hline \hline
  W/o & W/o & 73.41$\pm$1.18 & 76.66$\pm$1.02 & 74.67$\pm$5.76 & 71.89$\pm$5.91 \\
  W/o & With & 74.39$\pm$1.21 &\bf 77.90$\pm$0.90 & 67.78$\pm$1.11 & \bf 82.43$\pm$1.35 \\
  With & W/o & 73.74$\pm$1.39 & 77.15$\pm$3.80 & 68.10$\pm$7.20 & 81.15$\pm$6.56 \\
  With & With & \bf 75.34$\pm$0.47 & 77.03$\pm$1.85 &\bf 76.56$\pm$5.41 & 73.86$\pm$5.41 \\
\hline
\end{tabular}
}
\end{table}

% \subsection{Performance Discussion}
% One can observe from Table \ref{tab-performance} that HyperGALE achieves significant performance by justifying its intuition. The performance of HyperGALE is significant in both accuracy and AUC metrics {(\bf PLEASE INCLUDE TWO-SAMPLE T-TEST RESULTS HERE)}. Also, GCN is providing good sensitivity but has fluctuating performance with high standard deviation. Also, the specificity is higher for BrainNetTransformer \cite{kan2022brain} and HyperGALE performs on par with it.

% Thus, it can be observed that existing models based on transformers fail to answer the use of FC as tokens and the physical intuition behind attention. Graph-based models are more explainable but, still lack in acquiring higher-order relationships and thus lack in performance. Thus, our work HyperGALE processes the FC patterns to have better performance compared with existing state-of-the-art models by  emphasizing on functional associations that contribute to nuanced ASD patterns.

\subsection{Performance Discussion}
Table \ref{tab-performance} showcases that HyperGALE achieves notable performance, effectively justifying its design principles. The robustness of HyperGALE is confirmed in both accuracy and AUC metrics. A two-sample t-test comparing HyperGALE with HyperGraphGCN (a hypergraph-based baseline) reveals a significant difference in accuracy (t-Statistic: 3.40, p-Value: 0.0094), highlighting HyperGALE's superior performance in this regard. When compared with BrainNetTransformer, HyperGALE exhibits a highly significant improvement in accuracy (t-Statistic: 8.55, p-Value: 0.000027). Additionally, a significant enhancement is also observed in AUC (t-Statistic: 2.54, p-Value: 0.0345). These results highlight the efficacy of HyperGALE in handling functional connectivity (FC) patterns, particularly in Autism Spectrum Disorder (ASD) pattern identification.

While GCN demonstrates promising sensitivity, it suffers from fluctuating performance characterized by a high standard deviation. In terms of specificity, BrainNetTransformer excels, yet HyperGALE achieves comparable results. The superior performance of HyperGALE, especially in processing FC patterns, emphasizes its capability in delineating nuanced ASD patterns. Unlike transformer-based models that struggle with the utilization of FC as tokens and the physical understanding of attention mechanisms, HyperGALE adeptly processes these FC patterns, thereby capturing functional associations crucial in ASD diagnostics. This approach not only bolsters the model's accuracy and AUC but also enriches the understanding of ASD through cutting-edge machine learning techniques.

% One can observe from Table \ref{tab-performance} that HyperGALE achieves significant performance by justifying its intuition in acquiring information from higher-order proximities, attention networks, and adaptive readout mechanisms. % The performance of HyperGALE is significant in both accuracy and AUC metrics. Also, one can see that GCN is providing good sensitivity but has fluctuating performance (higher std). Also, the specificity is higher for BrainNetTransformer \cite{kan2022brain} and HyperGALE performs on par with it. % It can be observed that existing models based on transformers fail to answer the use of FC as tokens and the physical intuition behind attention. Graph-based models are more explainable since the learning is based on the underlying aggregation process modeled by the convolution operation. Further, we model the brain as a hypergraph and try to capture higher-order relations between \emph{functionally distant nodes} of the brain graph, which are not possible in generic graphs. This eventually guides in acquiring rich and invariant features that are sufficient to classify the subjects with ASD. Thus, our work HyperGALE processes the FC patterns in navigable terrain, guiding us through the complex neural landscape and ultimately emphasizing the functional associations that contribute to nuanced ASD patterns.

\subsection{Generalizability}

\begin{figure}[!t]
\centering
\includegraphics[width=1.0\linewidth]{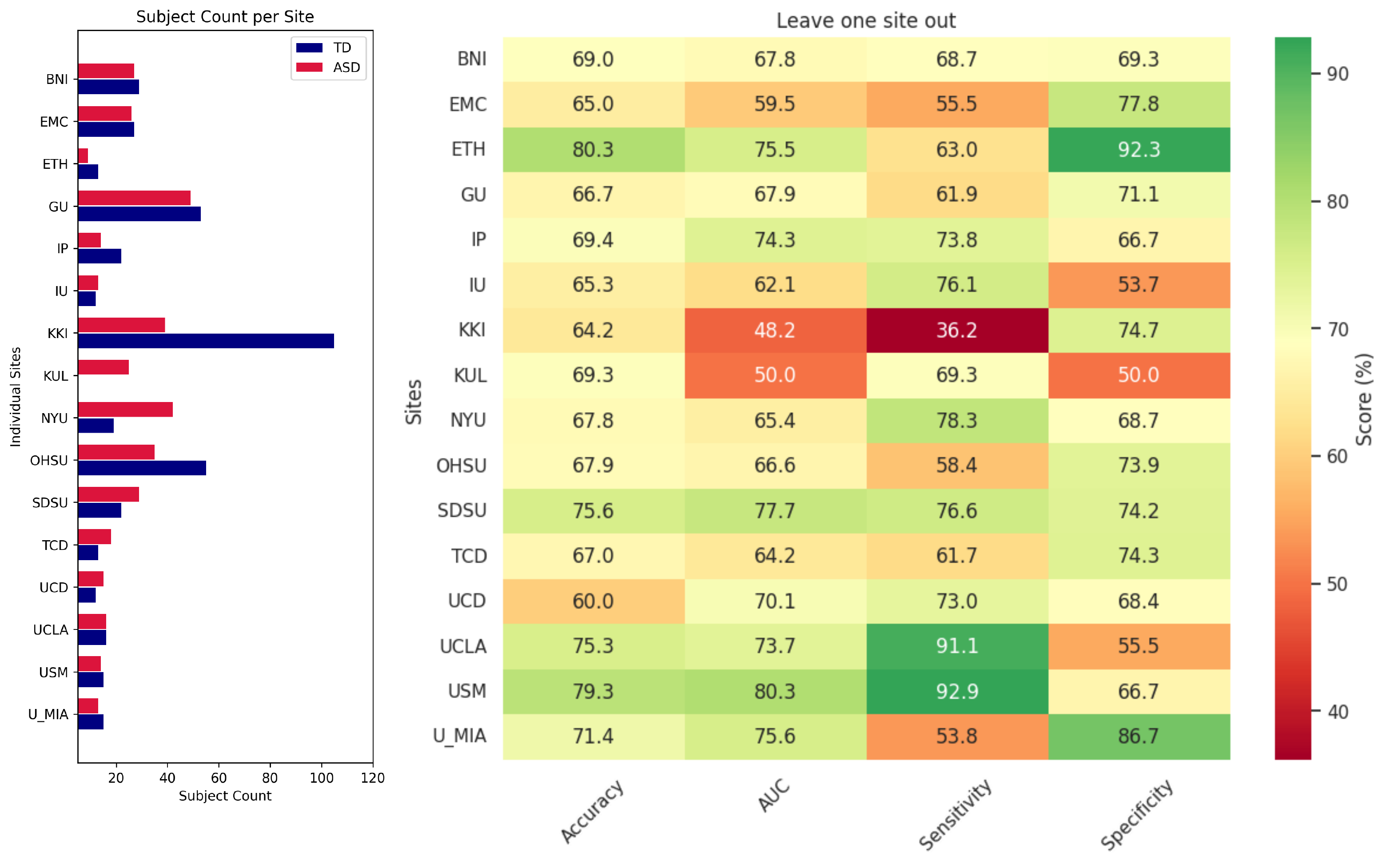}
\caption{\textbf{Prediction performance across sites using a leave-one-site-out strategy}. The count of ASD and TD subjects from various sites are shown on (left) and various performance metrics are shown for each site on (right). Chance level performance is at 50\%. Despite the challenges with site-specific variations in the number of samples, our model is still able to demonstrate creditable between-site generalization performance, comparable to the results in Table~\ref{tab-performance}.}
%Notably, at Site KUL, the absence of AUC and Sensitivity data, marked as `NA', is due to the lack of TD samples.}
\label{fig:leave-one-site-out}
\end{figure}

Our study employed a leave-one-site-out analysis across 16 different sites (refer Fig. \ref{fig:leave-one-site-out}) in the ABIDE-II dataset to evaluate the HyperGALE model's generalizability. This approach rigorously tested the model's performance in diverse settings, revealing insights into its applicability across various clinical environments. The analysis in Fig. \ref{fig:leave-one-site-out} demonstrates varied performance metrics - across sites. 
% Notably, site KUL presented a unique case with only ASD subjects, rendering AUC and specificity calculations impractical.
% For statistical consistency and to maintain dataset integrity, these metrics were replaced with a nominal value of 0.5, representing a neutral baseline indicative of random classification.

The results underscore the challenge of generalizing models in multi-site studies, highlighting the influence of site-specific factors such as data acquisition protocols and demographic variations. Performance discrepancies across sites suggest the need for further model enhancements, potentially through site-adaptive methodologies.

% \begin{figure}[!t]
%     \centering
%     \includegraphics[width=1.2\linewidth]{leave_one_site_out.png}
%     \caption{This figure presents the performance metrics obtained using a leave-one-site-out strategy on the ABIDE-II dataset across various sites. The absence of data (indicated by empty cells) at Site KUL for AUC and Sensitivity reflects the lack of typically developing (TD) samples at this location. Consequently, AUC and specificity calculations were not feasible for Site KUL, resulting in these values being designated as 'NA'.  
%     % In site KUL, only ASD subjects were present, hence AUC and specificity cannot be calculated. Hence, these values are replaced by 0.5.
%     }
%     \label{fig:leave-one-site-out}
% \end{figure}

% \begin{figure}[!t]
%     \centering
%     \includegraphics[width=0.99\linewidth]{Sites.png}
%     \caption{The figure provides how the ABDIE-II dataset was extracted from various sites and their contribution towards ASD and TD subjects.}
%     \label{fig-sites}
% \end{figure}

\subsection{Gated Attention and Hyperedge Discussion}

Table \ref{tab-GA-HE} demonstrates the impact of incorporating Learned Hyperedges (HE) and the Gated Attention (GA) module in hypergraph models, assessed under various configurations with and without these modules. One can observe that, with GA, the model is performing well in AUC and specificity metrics even in the absence of learnable HE. But, learnable HE and GA together actually obliges the model to have higher accuracy score and sensitivity. The sensitivity is helpful to predict the presence the ASD. Hence, both of these act as tandem and pivotal in identifying key ROI embeddings, thereby focusing on the most significant regions for predicting ASD. Furthermore, as GA highlights significant ROIs, and learned HE assign varying weights to different networks in brain parcellation, learning in this way helps the model to focus on distinct ROIs to provide significant performance. 

The significance of learned HE and the GA module extends beyond performance; they are also crucial for enhancing the model's interpretability in identifying biomarkers, a topic we will explore in detail in Section \ref{sec-interpret}.

% The findings reveal that the inclusion of the GA module significantly enhances the model's performance. This is evident from the improved metrics in the configurations where the GA module is present. The GA module is pivotal in identifying key ROI embeddings, thereby focusing on the most significant regions for predicting ASD. Furthermore, as GA highlights significant ROIs, Learned Hyperedges assign varying weights to different networks in brain parcellation, fine-tuning the model's focus and accuracy, hence, the combination of both, constituting the HyperGALE model, yields the most favourable outcomes.

% The importance of Learned Hyperedges and the GA module extends beyond performance; they are also crucial for enhancing the model's interpretability in identifying biomarkers, a topic we will explore in detail in Section VI.

\subsection{Ablation Study}

\begin{figure}[!t]
    \centering
    \includegraphics[width=1.0\linewidth]{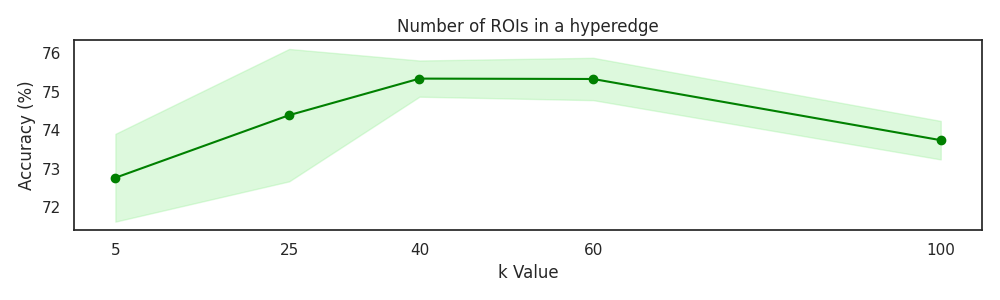}
    \caption{\textbf{Changes in prediction accuracy with respect to change in number of ROIs in a hyperedge denoted by k}. The optimal number of ROIs was found at $k$ = 40.}
    \label{fig:k-val}
\end{figure}

\paragraph{Number of ROIs in each Hyperedge} In our hypergraph convolution approach, hyperedges are formed by grouping ROIs based on connectivity, with the $k$-NN algorithm determining the number of ROIs in each hyperedge. The choice of the hyperparameter $k$, which dictates the ROI count per hyperedge, is crucial as it influences the overlap of hyperedges, impacting the identification of active brain regions in ASD and TD subjects \cite{lee2021hyperedges}.  A lower $k$ value might exclude vital information by limiting the ROIs per hyperedge, whereas a higher $k$ risks adding non-significant connections and noise. Our empirical analysis as seen in Fig. \ref{fig:k-val}, determined 40 to be the optimal ROI count in a hyperedge, balancing comprehensive coverage and minimal noise, crucial for accurate brain region representation in ASD and TD subjects.

% Hyperedges are created before doing hypegraph convolution by grouping together all the ROIs which are in proximity based on connectivity. For every ROI, a hyperedge is created by applying KNN algorithm. Hyperparameter k decides how many ROIs will be present in each hyperedge. \cite{lee2021hyperedges} highlights the importance of choosing the number of nodes in a hyperedge as as it can influence the overlap hyperedges. The overlap in brain ROIs can be significant in deciding which regions of brain are more active in Austic and Typically developing subjects. 

% Selecting a lower k leads to fewer ROIs being included in each hyperedge. This approach may result in excluding critical information from the brain's connectome structure, as only a few edges are chosen for the hypergraph convolution. Conversely, a higher k incorporates more ROIs into each hyperedge. While this can ensure a more comprehensive representation, it also risks including numerous non-significant connections, potentially introducing noise into the analysis. 

% Our empirical analysis, guided by classification accuracy and the representation of brain regions in ASD and TD subjects, identified 40 as the optimal number of ROIs in a hyperedge. This ensures sufficient coverage of the connectome structure without overwhelming the model with irrelevant connections.

\begin{figure}[!t]
    \centering
    \includegraphics[width=1.0\linewidth]{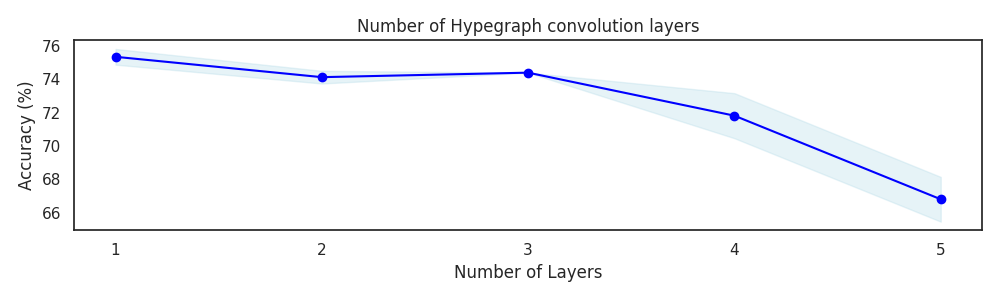}
    \caption{\textbf{Changes in prediction accuracy with respect to change in the number of hypergraph convolution layers.} A single-layer hypergraph convolution is found to be optimal.}
    \label{fig:L-val}
\end{figure}

\paragraph{Number of Layers} Our analysis shows that a single-layer hypergraph convolution is optimal for brain connectome analysis, as evidenced in Fig. \ref{fig:L-val}. An increase in the number of layers led to a decline in performance, a phenomenon consistent with over-smoothing issues in graph convolutional networks (GCNs). This issue, as discussed in \cite{li2018deeper, chen2020measuring}, involves feature homogenization across layers, reducing the model's ability to distinguish unique patterns.

\begin{table}[t!]
\caption{The Table illustrates the performance of various readout mechanisms on our proposed HyperGALE model.}\label{tab-readouts}
% \scalebox{1.05}{
\begin{center}
\begin{tabular}{ l|cccc } 
\hline
\multirow{2}{5em}{\textbf{Methods}} & \multicolumn{4}{c}{\textbf{Performance Metrics}} \\ \cline{2-5}
    & Accuracy & AUC & Sensitivity & Specificity \\ \hline \hline
  MLP  &  \bf 75.34 &	\bf 77.03 &	\bf 68.10	&81.15  \\
  Set Transformer & 67.36 &	65.95 &	41.03 &	77.14 \\
  
  Janossy & 62.96 & 58.08 &	63.16 &	62.79 \\
  
  Max &51.12 &	54.14& 23.00 &	\bf 97.30\\
  
  Mean &48.17& 50.36& 10.00	& 94.59\\
\hline
\end{tabular}
\end{center}
% }
\end{table}

The brain connectome's structure, mainly consisting of ROIs one hop away from each other, supports our use of a single-layer approach. This structure allows for effective aggregation of essential information in the initial layer, making additional layers redundant. While various strategies have been proposed to address over-smoothing in deeper GNNs \cite{liu2020towards, li2021deepgcns}, the architectural specifics of our dataset make these approaches less relevant, as our single-layer model adequately captures the necessary connectivity information.

% Our analysis, as illustrated in Fig. \ref{fig:L-val}, demonstrates that the optimal performance in hypergraph convolution for brain connectome analysis is achieved with a single layer. We observed a decline in results with an increase in the number of layers. This phenomenon aligns with the known issue of over-smoothing in graph convolutional networks (GCNs), a challenge extensively discussed in the literature. As \cite{li2018deeper, chen2020measuring} have noted, additional layers in GCNs can lead to homogenization of features across the graph, diminishing the model's ability to capture distinctive patterns.

% The brain connectome's structure, characterized by ROIs predominantly one hop away from each other, further substantiates our finding. This proximity implies that essential information is effectively aggregated in the initial layer of hypergraph convolution, negating the need for multiple layers. While various approaches have been proposed to mitigate over-smoothing in deeper GNNs, as suggested by \cite{liu2020towards, li2021deepgcns}, our study's context renders these solutions less pertinent. The single-layer model efficiently captures the necessary connectivity information within the brain connectome, aligning with the specific architectural characteristics of our dataset.

\paragraph{Readout Mechanisms}

% As mentioned earlier, the readouts such as, $\textsc{max} (\cdot) $ and $\textsc{mean} (\cdot)$ underfit while learning on brain hypergraphs and the conclusive evidence can be seen in Table. \ref{tab-readouts}. 

% Buterez \emph{et al.} \cite{buterez2022graph} emphasized that all readouts cannot be generalized to every kind of graph structure, and for some cases, permutation non-invariant can work better than invariant readouts. Also, to every instance and permutation invariant property is not always a necessity and since brain graphs are fully connected we should expect even permutation variant readouts to give correct embeddings. The $\textsc{SetTransformer}$ readout was achieved without the positional encoding to preserve the permutation invariance property of graph readouts. Murphy \emph{et al.}\cite{murphy2018janossy} propose  $\textsc{Janossy}$, a permutation-invariant readout function as the average of a permutation-sensitive function applied to all re-orderings of the input sequence. 

% For brain hypergraphs as well, MLP readout gives significantly better embeddings compared to standard and adaptive permutation invariant methods. Thus,  we have utilised  $\textsc{Mlp}$ readouts as they comply with our framework producing significant representations \cite{alon2020bottleneck}. Alon \emph{et al.} \cite{alon2020bottleneck} prone to have significant improvement in performance and it justifies our use of  $\textsc{Mlp}$ readouts. 

Table \ref{tab-readouts} demonstrates that standard readouts like $\textsc{max} (\cdot)$ and $\textsc{mean} (\cdot)$ are less effective for brain hypergraphs. Supporting this, Buterez \emph{et al.} \cite{buterez2022graph} note that not all readouts suit every graph structure, especially in fully connected graphs like brain hypergraphs where permutation non-invariant readouts can be more effective.

The $\textsc{SetTransformer}$ readout, used without positional encoding to preserve permutation invariance, contrasts with the permutation-invariant $\textsc{Janossy}$ readout suggested by Murphy \emph{et al.}\cite{murphy2018janossy}. However, in our study, MLP readouts provided superior embeddings for brain hypergraphs, aligning with findings by Alon \emph{et al.} \cite{alon2020bottleneck} that demonstrate their effectiveness in similar contexts. Hence, our framework utilized $\textsc{MLP}$ readouts for optimal representation.


\section{Interpretability Discussion}\label{sec-interpret}

Our analysis of brain connectome interactions unveils distinct neural patterns between ASD and TD subjects. Utilizing the Gated Attention module and learned hyperedge weights, we identified pivotal ROIs and observed how their interactions differ between ASD and TD individuals.

\begin{figure}[!t]
    \centering    \includegraphics[width=0.48\textwidth]{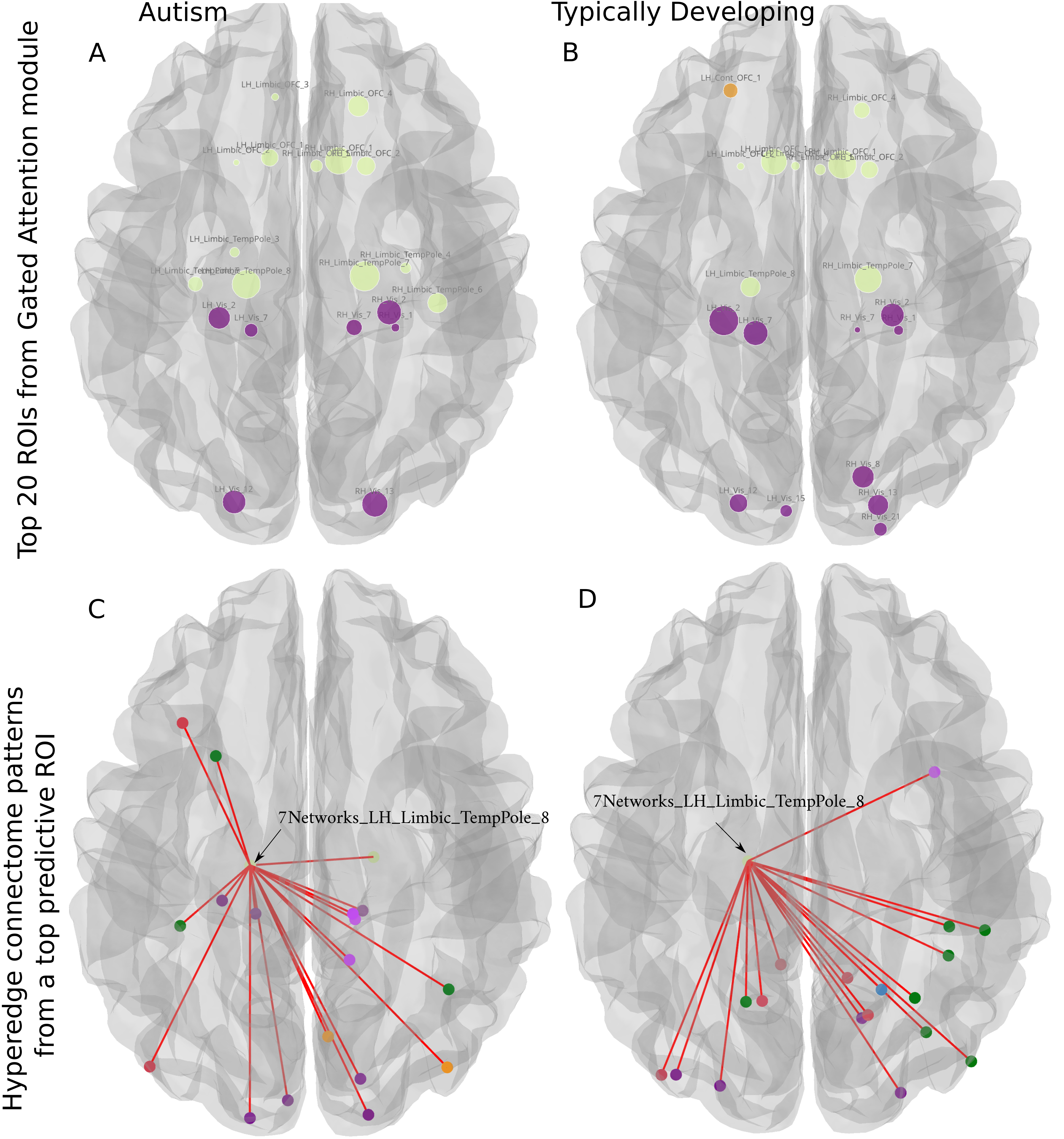}
    \caption{ \textbf{The interpretation of important ROIs and their connectivity obtained from GA module and learned hyperedge weights, respectively}. (A) and (B) highlight the top 20 most important ROIs that emerged in ASD and TD subjects. (C) and (D) depict the \emph{discriminative hyperedge connectivity pattern} of a \emph{common} ROI in ASD and TD using dominant hyperedge based on $\Tilde{W}$. %The patterns of connectivity are different across ASD and TD, although the node is common and important in both.
    % Interpretation of the Brain Hypergraphs with gated attention. (A) and (B) highlight the top $20$ most important ROIs that emerged in ASD and TD subjects. Whereas (C) and (D) show the discriminative hyperedge connectivity of common ROI in ASD and TD using dominant hyperedge based on $\Tilde{W}$.
    }
    \label{fig:interpretation}
\end{figure}

While Figs \ref{fig:interpretation} (A) and (B) reveal a significant overlap in the top ROIs across both ASD and TD groups, it is the distinct connections among regions of these two groups that provide deeper insights and distinguish the two groups. In other words, despite sharing common ROIs, the way these regions interact within the neural network varies considerably between ASD and TD, reflecting the unique neural connectivity characteristics of ASD.

We identified the top 50 ROIs and categorized them into broader functional networks, each reflecting distinct aspects of the ASD classification. In the following we highlight the differences observed in the functional networks between ASD and TD. These include, Visual Processing Networks, which highlight atypical visual processing patterns in ASD \cite{lombardo2019default}. The Limbic System indicates altered emotional and social processing through higher connectivity with the Default Mode Network (DMN) \cite{bigler2003temporal}. Executive Function and Attention Networks reveal disparities related to executive functioning and attention regulation challenges in ASD; and the Default Mode Network, where atypical connectivity patterns impact areas like theory of mind and self-awareness \cite{bigler2003temporal}.

% We broadly categorized these important ROIs we obtained into networks based on their cognitive and neural roles:

% \begin{itemize}
%     \item Visual Processing Networks: Highlights differences in sensory processing, with ASD showing atypical visual processing patterns \cite{lombardo2019default}.
%     \item Limbic System: More pronounced connectivity with the Default Mode Network (DMN) in ASD points to altered emotional and social processing, aligning with the challenges in empathy and social interaction commonly observed in ASD \cite{bigler2003temporal}.
%     \item Executive Function and Attention Networks: Disparities in these networks underline the executive functioning and attention regulation challenges in ASD.
%     \item Default Mode Network: Atypical connectivity patterns in ASD impact critical areas such as theory of mind and self-awareness \cite{bigler2003temporal}.
% \end{itemize}

While Figs. \ref{fig:interpretation} (A) and (B) indicate a substantial overlap in the top ROIs across both ASD and TD groups, it is their distinctive connectivity patterns that yield profound insights. This is particularly evident in the Limbic System, comprising of regions like Limbic TempPole and Limbic OFC. In ASD, we observe higher connectivity of these limbic regions with the DMN, as opposed to the balanced connections seen in TD. This higher linkage in ASD, as highlighted in Fig \ref{fig:interpretation} (C) versus (D), is indicative of altered connectivity patterns in areas crucial for emotional processing and social interaction \cite{bigler2003temporal}. Similarly, in visual regions (RH Vis, LH Vis), we observe increased connectivity to DMN regions in ASD subjects \cite{lombardo2019default}, while TD had connections primarily within visual and attention networks.

These observations improve our understanding of ASD's neural dynamics. The Limbic System's enhanced DMN connectivity in ASD might underlie social and emotional processing challenges, while the varied connectivity in visual networks points to broader sensory and cognitive implications. Overall, our study extends beyond identifying shared ROIs, and discovering complex variations in neural interactions.% that are essential for understanding ASD's neural framework.

% These findings are critical as they align with existing literature suggesting atypical limbic system functionality in ASD. The Limbic System's stronger connections with the DMN in ASD could reflect the condition's characteristic challenges in social cognition and emotional regulation. Also, visual networks ASD subjects shows more connections outside the primary visual processing areas, involving regions related to self-referential thought (DMN). Our results extend beyond just identifying shared ROIs; they reveal the intricate variations in neural interactions.

% shorter conclusion
% \section{Conclusion}
% In this paper, we present HyperGALE, an advanced hypergraph convolutional network with gated attention, tailored for ASD classification. Our evaluations on the ABIDE-II dataset demonstrate HyperGALE's superior performance over traditional machine learning and neural network approaches. By effectively capturing high-order graph complexities and introducing learnable hyperedge weights, HyperGALE offers a refined understanding of brain network dynamics by offering the regions to look for to identify ASD, and how does connectivity of these regions vary from ASD to TD subjects. This information can not only help clinically detect ASD but open avenues for research in neuroimaging and intervention strategies. 

\section{Conclusion}
In this paper, we introduced HyperGALE, a novel hypergraph convolutional network enhanced with gated attention, specifically designed for ASD classification. Our comprehensive evaluation on the ABIDE-II dataset highlights HyperGALE's outstanding performance, surpassing ML and graph-based approaches. HyperGALE's capability to effectively capture complex higher-order graph intricacies and utilize learnable hyperedge weights has led to a better understanding of brain network dynamics. HyperGALE identifies critical ROIs for ASD and delineates how their connectivity patterns differ from those in TD subjects. This, not only promises improvements in clinical detection but also opens new avenues for research in neuroimaging and the development of targeted intervention strategies.

\bibliographystyle{ieeetr}
\bibliography{main}

\vspace{12pt}
% \color{red}
% IEEE conference templates contain guidance text for composing and formatting conference papers. Please ensure that all template text is removed from your conference paper prior to submission to the conference. Failure to remove the template text from your paper may result in your paper not being published.

\end{document}